%% file: main.tex
\documentclass[runningheads]{llncs}

\usepackage{eccv}

\usepackage{eccvabbrv}

\usepackage{graphicx}
\usepackage{booktabs}

\usepackage[accsupp]{axessibility}  

\usepackage[pagebackref,breaklinks,colorlinks,citecolor=eccvblue]{hyperref}
\usepackage{makecell}
\usepackage{multirow} 
\usepackage{wrapfig} 
\usepackage{caption}
\usepackage{orcidlink}
\usepackage{xcolor} 
\definecolor{darkgreen}{rgb}{0,0.5,0}
\newcommand{\zj}[1]{\textcolor{black}{#1}}

\newcommand*{\rom}[1]{\uppercase\expandafter{\romannumeral #1\relax}}

\begin{document}

\title{Understanding and Mitigating Human-Labelling Errors in Supervised Contrastive Learning} 


\author{Zijun Long
\and
Lipeng Zhuang
\and
George Killick
\and
Richard McCreadie
\and \\
Gerardo Aragon-Camarasa
\and
Paul Henderson
}


\institute{University of Glasgow, Glasgow, United Kingdom\\
\email{z.long.2@research.gla.ac.uk,
\{lipeng.zhuang, george.killick,\\richard.mccreadie, gerardo.aragoncamarasa, paul.henderson\}@glasgow.ac.uk}}

\maketitle

\input{sec/0_abstract}    
\input{sec/1_intro}

\input{sec/related_work}
\input{sec/cl_setup}
\input{sec/label_analysis}

\input{sec/method}

\input{sec/exp}
\input{sec/conclusion}

\clearpage  

%
%
\bibliographystyle{splncs04}
\bibliography{main}
\end{document}

%% file: sec/0_abstract.tex
\begin{abstract}
Human-annotated vision datasets inevitably contain a fraction of human-mislabelled examples. While the detrimental effects of such mislabelling on supervised learning are well-researched, their influence on Supervised Contrastive Learning (SCL) remains largely unexplored. 
In this paper, we show that human-labelling errors not only differ significantly from synthetic label errors, but also pose unique challenges in SCL, different to those in traditional supervised learning methods.
Specifically, our results indicate they adversely impact the learning process in the \(\sim\!\!99\%\) of cases when they occur as false positive samples.
\zj{Existing noise-mitigating methods primarily focus on synthetic label errors and tackle the unrealistic setting of very high synthetic noise rates (40--80\%), but they often underperform on common image datasets due to overfitting. To address this issue, we introduce a novel \zj{SCL objective with robustness to human-labelling errors}, SCL-RHE. SCL-RHE is designed to mitigate the effects of real-world mislabelled examples, typically characterized by much lower noise rates ($<5\%$).}
We demonstrate that SCL-RHE consistently outperforms state-of-the-art representation learning and noise-mitigating methods across various vision benchmarks, by offering improved resilience against human-labelling errors.

  \keywords{Contrastive Learning \and Label Noise \and Image Classification}
\end{abstract}

%% file: sec/1_intro.tex
\section{Introduction}

\label{sec:intro}

Contrastive methods achieve excellent performance on self-supervised learning \cite{RN89, npairloss, DBLP:journals/corr/abs-1810-06951}. They produce latent representations that excel at many downstream tasks, from image recognition and object detection to visual tracking and text matching \cite{suconlan,Tan2022DomainGF}. \textit{Supervised} contrastive learning (SCL) utilizes label information to improve representation learning, encouraging closer distances between same-class samples (positive pairs) and greater distances for different-class samples (negative pairs). SCL outperforms traditional methods for pre-training that employ a cross-entropy loss \cite{suconlan, SSCL, RN81}.

However, the effectiveness of SCL depends on the correctness of the labels used for identifying image pairs to contrast. Human-labelling errors introduce erroneous positive and negative pairings, compromising the integrity of learned representations \cite{SSCL}. Even widely-used datasets exhibit significant numbers of mislabeled images—for example, the ImageNet-1K validation set has 5.83\% of images wrongly labeled \cite{DBLP:conf/nips/NorthcuttAM21}.
\zj{The impact of noisy labels on supervised learning has been extensively researched \cite{DBLP:journals/tnn/SongKPSL23,DBLP:journals/corr/abs-2011-04406,DBLP:conf/iccS/DubelWN23}.} However, the extent and manner in which human-labelling errors influence SCL remains under-explored. As SCL emerges as a compelling alternative to the cross-entropy loss, it becomes increasingly important to investigate human-labelling errors specifically in the context of SCL.

\begin{figure*}[t]
  \centering
   \includegraphics[width=1.0\linewidth]{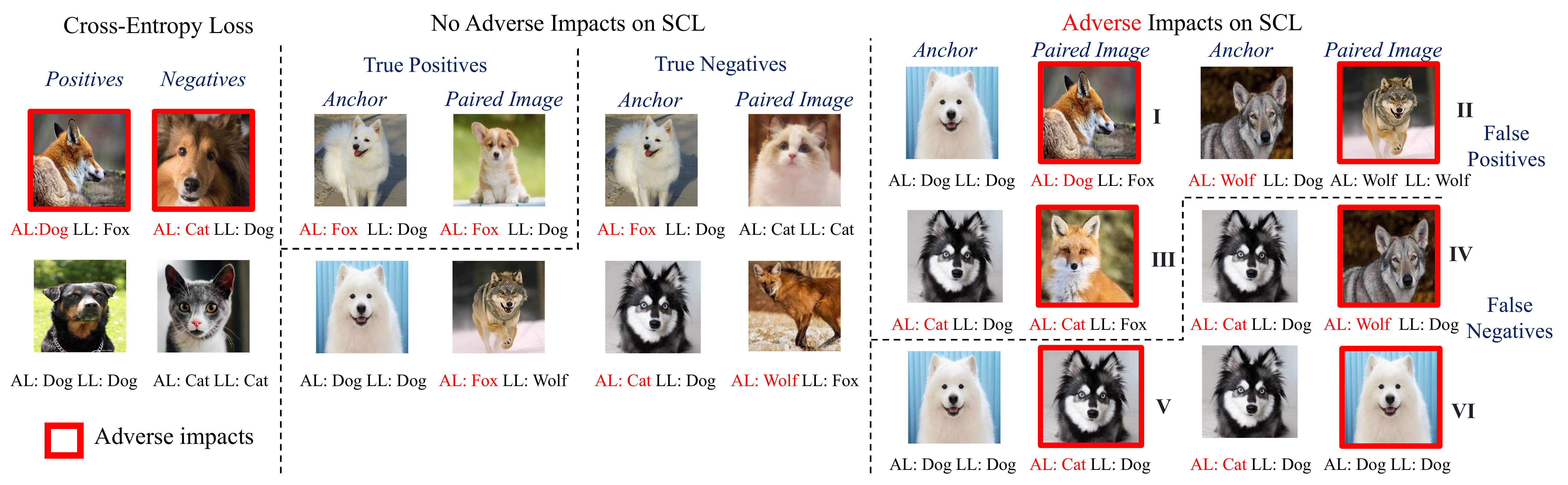}

   \caption{Comparison between impacts of labelling errors on different learning approaches. AL represents `Assigned Label' and LL represents `Latent Label'. Those marked red in AL represent human-labelling errors. 
   \zj{It is important to note that as long as a pair shares the same latent label, there are no adverse impacts on positive pairs. Similarly, if the latent labels differ, negative pairs remain unaffected.}
   } 
   \label{fig:CL_compare}
\end{figure*}

In this paper, we first analyze how huma-labelling errors affect SCL and how they differ from supervised learning with cross-entropy loss. As shown on the left side of Fig.~\ref{fig:CL_compare} and detailed in Sec.~\ref{sec:analysis_on_truepair}, labelling errors, regardless of being positive or negative, negatively impact supervised learning with cross-entropy loss during training. Existing noise-mitigation methods, based on SCL or cross-entropy loss approaches \cite{SSCL,huang2023twin}, aim to eliminate labelling errors from training samples. However, we argue that such strategies, often detrimental to SCL, compromise the quality of learned representations. The middle of Fig.~\ref{fig:CL_compare} shows that labelling errors do not always adversely affect SCL, with their removal potentially reducing training sample size and lowering overall performance. Unlike in cross-entropy methods, labelling errors exhibit more complex dynamics in how they affect learning signals in SCL. Both correctly and incorrectly labeled images can generate correct or incorrect learning signals, depending on the images they are paired with. Our analysis in Sec.~\ref{sec:analysis_on_truepair} reveals that nearly 99\% \zj{of incorrect learning signals in SCL} are due to mislabeled positive samples. This highlights the need for a tailored SCL strategy that effectively addresses human-labelling errors without sacrificing performance.

\zj{Although human-labelling errors are prevalent in image datasets, existing noise-mitigation methods predominantly focus on synthetic labelling errors, showing good performance when noise is intentionally added at levels ranging from 40\% to 80\% \cite{DBLP:conf/icml/MaWHZEXWB18,DBLP:conf/icml/ZhengWG0MC20,DBLP:conf/aaai/ChenYCZH21,SSCL,huang2023twin}. Yet, in scenarios with realistic human-labelling errors, such as those found in ImageNet-1K (with a noise rate of approximately 5.83\% \cite{DBLP:conf/nips/NorthcuttAM21}), these methods underperform, especially against cross-entropy based methods, primarily due to overfitting \cite{rethinkgeneralization,DBLP:journals/tnn/SongKPSL23,DBLP:journals/corr/abs-2011-04406,DBLP:conf/iccS/DubelWN23, DBLP:conf/nips/NorthcuttAM21}. For example, methods that rely on assigning confidence values to pairs and prioritizing learning from those deemed confident \cite{SSCL,huang2023twin} risk overfitting to incorrect labels and neglect a significant portion of training data, an issue exacerbated in datasets with many similar classes \cite{DBLP:journals/tnn/SongKPSL23,DBLP:conf/cvpr/YaoSZS00T21}. Additionally, existing noise-mitigating methods also introduce significant computational complexity and overhead. For example, Sel-CL \cite{SSCL} is challenging to apply on large datasets since it uses the $k$-NN algorithm to create pseudo-labels.
}

\zj{Importantly, in Sec.~\ref{sec:similarity_of_mis}, we reveal  significant differences between the distributions of synthetic and human-labelling errors, emphasizing the need for tailored mitigation strategies. Our empirical analysis indicates that human-labelling errors often stem from high visual similarity between assigned and actual classes, leading to notable overlaps in the representation distributions of correctly and incorrectly labeled samples. This indicates that human-mislabeled samples are often indistinguishable in the representation space to samples sharing the same assigned label; this contrasts with synthetic label errors that are more arbitrary.  Thus, there is an urgent need for methods that not only mitigate the impact of human-labelling errors in SCL on widely-used human-annotated datasets like ImageNet-1K but also preserve computational efficiency. }


\zj{Based on our analysis and existing research gaps}, we propose a novel SCL objective with robustness to human-labelling errors that emphasises true positives, which come from the same latent class but are are far apart in feature space (Sec.~\ref{sec:method}). In contrast, existing \zj{noise-mitigating methods tuned for synthetic noise} \cite{DBLP:conf/icml/ArazoOAOM19,DBLP:conf/sisap/Houle17} typically assign greater weight to confident pairs that are closely positioned, despite the high likelihood of these pairs being false positives. Furthermore, based on the established benefits of utilizing `hard' negative samples \cite{DBLP:journals/corr/SongXJS15, DBLP:journals/corr/SchroffKP15,DBLP:conf/iclr/RobinsonCSJ21}, we hypothesise that \textit{true positives originating from the same latent class, yet positioned distantly, are important for enhancing the quality of learned representations}. 

In summary, our main contributions are:
\begin{itemize}
\item We present an in-depth analysis elucidating the manner and impact of human-mislabeled samples on SCL and offering strategies to effectively mitigate them (Sec.~\ref{sec:label_analysis}).
\zj{\item  We introduce a novel SCL objective, \textit{SCL-RHE}, distinguished as the first to specifically address human-labelling errors by focusing on false positives within SCL (Sec.~\ref{sec:method}). Unlike previous works \cite{DBLP:conf/icml/ZhengWG0MC20,DBLP:conf/aaai/ChenYCZH21,SSCL,huang2023twin}, SCL-RHE not only demonstrates state-of-the-art performance on widely used image datasets, effectively avoiding overfitting, but also maintains efficiency by not adding extra computational overhead.
\item We show the broad applicability of the proposed SCL-RHE objective across two distinct learning scenarios: training from scratch and transfer learning. SCL-RHE outperforms previous state-of-the-art SCL and noise-mitigation methods, achieving higher Top-1 accuracy on ten tested datasets, as detailed in Sec.~\ref{sec:exp}, in both learning scenarios. SCL-RHE sets a new state-of-the-art for base models (with 88M parameters) on ImageNet-1K.}

\end{itemize}

%% file: sec/related_work.tex
\section{Related Work}

\zj{Human-labelling errors} are prevalent in many datasets used for supervised learning, especially for large datasets where eliminating errors is impractical \cite{DBLP:conf/nips/NorthcuttAM21,DBLP:journals/corr/abs-2208-08464}. Mislabeled examples can lead to overfitting in models, with larger models being more susceptible \cite{rethinkgeneralization,DBLP:journals/tnn/SongKPSL23,DBLP:journals/corr/abs-2011-04406,DBLP:conf/iccS/DubelWN23, DBLP:conf/nips/NorthcuttAM21}. Robust learning from noisy labels is thus crucial for improving generalization. Methods include estimating noise transition matrices \cite{holviewtransmat, transmat2, dual-t-transmatrix}, regularization \cite{DBLP:journals/corr/abs-1710-09412, RegularizeNoise, DBLP:conf/eccv/JenniF18}, and sample re-weighting \cite{DBLP:conf/icml/RenZYU18,DBLP:journals/tnn/WangLT18}. The influence of human-labelling errors in supervised contrastive learning (SCL) is less explored; we address this by \zj{developing dedicated methods to mitigate human-labelling errors.}

Contrastive learning, particularly in unsupervised visual representation learning, has evolved significantly \cite{becker1992self, RN89, mocov2, barlowtwins, byol, swav}. However, the absence of label information can result in positive samples in negative pairs, potentially leading to detrimental effects on the representations learned. SCL leverages labeled data to construct positive and negative pairs based on semantic concepts of interest (e.g.~object categories). It ensures that semantically related points are attracted to each other in the embedding space.  Khosla et al.~\cite{RN81} introduce a SCL objective inspired by InfoNCE\cite{Gutmann2010NoisecontrastiveEA,npairloss,oord2018representation}, which can be considered a supervised extension of previous contrastive objectives---e.g. triplet loss \cite{DBLP:journals/corr/SchroffKP15} and N-pairs loss \cite{npairloss}). Despite its efficacy, the impact of label noise and the importance of hard sample mining in SCL are often overlooked. We build on their work by devising a strategy that not only \zj{mitigates the impact of human-labelling errors but also enhances the performance of the SCL objective.}

\zj{Previous noise-mitigation works in SCL primarily target synthetic labelling errors, often excluding human-mislabeled samples through techniques like specialized selection pipelines \cite{DBLP:journals/corr/SongXJS15, DBLP:journals/corr/SchroffKP15} and bilevel optimization \cite{DBLP:conf/eccv/JenniF18}. Recently, Sel-CL \cite{SSCL} introduced a non-linear projection head for intra-sample similarity analysis to identify confident samples, and TCL \cite{huang2023twin} employs a Gaussian mixture model with entropy-regularized cross-supervision. Despite these advances, such methods risk overfitting on these confident pairs, particularly in the presence of human-labelling errors, reducing their effectiveness in real-world scenarios as shown in our analysis (Sec.~\ref{sec:analysis_on_truepair}). They also focus on artificially noisy datasets with synthetic noise rates up to 80\%, an unrealistic scenario in practice. Our method instead specifically targets human-labelling errors in common image datasets, optimizing SCL performance without excluding all of these errors.}

%% file: sec/cl_setup.tex
\section{\zj{Setup for Contrastive Learning}}

We begin by discussing the fundamentals of contrastive representation learning. Here, the objective is to contrast pairs of data points that are semantically similar (positive pairs) against those that are dissimilar (negative pairs). Mathematically, given a data distribution \( p(x) \) over \( \mathcal{X} \), the goal is to learn an embedding \( f: \mathcal{X} \rightarrow \mathbb{R}^{d} \) such that similar pairs \( (x, x^{+}) \) are close in the feature space, while dissimilar pairs \( (x, x^{-}) \) are more distant. In unsupervised learning, for each training datum \( x \), the selection of \( x^{+} \) and \( x^{-} \) is dependent on \( x \). Typically, one positive example \( x^{+} \) is generated through data augmentations, accompanied with \( N \) negative examples \( x^{-} \). The contrastive loss, named InfoNCE or the $N$-pair loss \cite{Gutmann2010NoisecontrastiveEA,npairloss,oord2018representation}, is then defined as
%
\begin{equation}
\resizebox{0.55\textwidth}{!}{$
\mathcal{L}_{\mathrm{NCE}} = \mathbb{E}_{\substack{x\\ x^{+}\\ \left\{x_i^{-}\right\}_{i=1}^N}}\left[-\log \frac{e^{f(x)^T f(x^{+})}}{e^{f(x)^T f(x^{+})}+\sum_{i=1}^N e^{f(x)^T f(x_i^{-})}}\right]
$}
\label{eq:infoloss}
\end{equation}
In Eq. \ref{eq:infoloss}, the expectation computes the average loss across all possible choices of positive and negative samples within the dataset. In practice, during a training iteration, one typically samples a mini-batch; then, for each data point in it (referred to as an `anchor'), a positive example is selected---usually an augmented version of the anchor or another instance of the same class---while the rest of the batch is treated as negative examples. This is under the assumption that within the batch, instances of different classes (i.e., all other samples except the positive pair) serve as negatives. Khosla \textit{et al.}~\cite{RN81} extended this concept to supervised contrastive learning, experimenting with two losses:

\begin{equation}
\resizebox{0.63\textwidth}{!}{$
\mathcal{L}_{\text {in }}^{ \text {sup} } =
\mathbb{E}_{\substack{x\\ \left\{x_k^{+}\right\}_{k=1}^K \\\left\{x_i^{-}\right\}_{i=1}^N}}
 -\log \left\{\frac{1}{|K|} \sum_{k=1}^K \frac{\exp \left(f(x)^T f\left(x_k^{+}\right)\right)}{\sum_{k=1}^{K} e^{f(x)^T f( x_{k}^{+})}+\sum_{i=1}^N e^{f(x)^T f(x_i^{-})}}\right\}
$}
\label{eq:supercon_in} 
\end{equation}

\begin{equation}
\resizebox{0.63\textwidth}{!}{$
\mathcal{L}_{\text {out }}^{\text {sup }} = 
\mathbb{E}_{\substack{x\\ \left\{x_k^{+}\right\}_{k=1}^K \\\left\{x_i^{-}\right\}_{i=1}^N}}
\frac{-1}{|K|} \sum_{k=1}^K \left[ 
    \log \left\{  
        \frac{
            \exp \left(f(x)^T f\left(x_k^{+}\right)\right)
        }{
            \sum_{k=1}^{K} e^{f(x)^T f( x_{k}^{+})}+\sum_{i=1}^N e^{f(x)^T f(x_i^{-})}
        }
    \right\}
    \right]
$}
\label{eq:supercon_out}
\end{equation}
Here $k$ indexes a set of $K$ positive samples, i.e.~images $x_k^+$ of the same class as $x$. We focus on improving these objectives, making them more robust to human-labelling errors in Sec.~\ref{sec:method}.

%% file: sec/label_analysis.tex
\section{\zj{Uniqueness of Human-labelling Errors and Their Impact on SCL}}
\label{sec:label_analysis}
\zj{In this section, we show that human-labelling errors and synthetic label errors exhibit distinct characteristics. Human-labelling errors arise from the high visual similarity between the sample and its assigned class, making it challenging for humans to differentiate them accurately. In contrast, synthetic label errors are generated randomly and lack this similarity \cite{SSCL,huang2023twin}. This distinction underlines the need for a method specifically tailored to address the unique challenges human-mislabelled samples pose to supervised contrastive learning (SCL). We further illustrate the specific impact of these errors on SCL, distinct from their effect on supervised learning with cross-entropy, by analyzing various scenarios of mislabelling within SCL and assessing their adverse effects.}

\paragraph{Definitions.}
We define the \textit{latent label} of an image as being its true category (e.g.~the latent label of an image of a cat would be `cat').
The term \textit{assigned label} refers to the class that a human annotator has assigned to an image (hopefully---but not always---matching the latent label).
Given a pair of images,
we define a \textit{false positive} as being when an annotator has erroneously grouped those images under the same assigned label, even though their latent labels are different.
A \textit{false negative} is when two images sharing the same latent label mistakenly have different assigned labels.
We define \textit{true positive} and \text{true negative} pairs analogously.
Lastly, we define \textit{easy positives} as pairs of images that share the same assigned labels and have highly similar embeddings.

\subsection{\zj{The Differences Between Human-Labelling Errors and Synthetic Label Errors}}
\label{sec:similarity_of_mis}

\begin{figure}[t]
    \centering
    \includegraphics[width=1\linewidth]{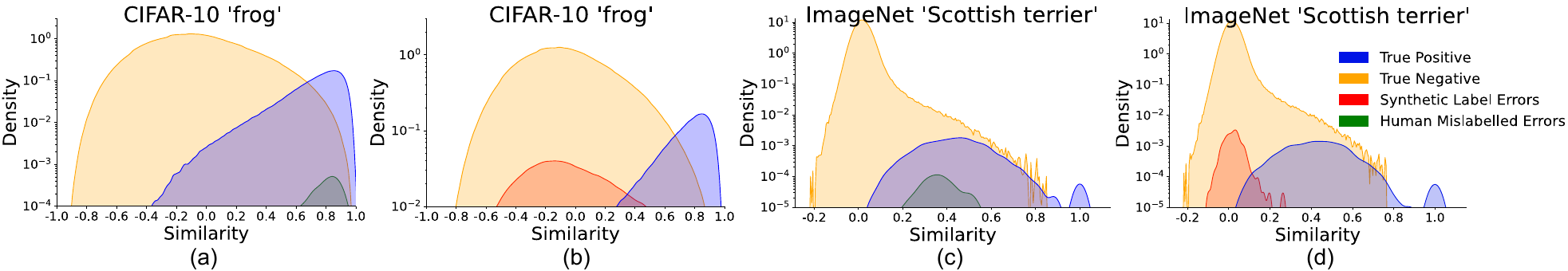}
    \caption{Figures (a) and (c) display the log-scaled distribution of cosine similarities for various pair types, including true positive pairs, true negative pairs, and human-labelling errors, on the CIFAR-10 and ImageNet-1k datasets, respectively. Conversely, figures (b) and (d) present analogous data, focusing instead on synthetic label errors.}
    \label{fig:similarity map}
\end{figure}

\zj{We begin our analysis with the question: \textit{What distinguishes human-labelling errors from synthetic label errors?} We pretrained ViT-base models on the CIFAR-10 and ImageNet-1k datasets individually, utilizing the SCL objective as defined in Eq.~\ref{eq:infoloss}. Then, we conduct a similarity analysis of the resulting features for different types of label errors within the context of contrastive learning. Specifically, we use the consensus among annotators from \cite{DBLP:conf/nips/NorthcuttAM21} to identify human-mislabelled samples, and synthetic errors are produced by randomly altering 20\% of the labels to different classes.} \zj{In Fig.~\ref{fig:similarity map}, we plot the similarity distributions for various pair types across two image classes, contrasting human-labelling errors with synthetic label errors. As we can see from Fig.~\ref{fig:similarity map}, there is a significant overlap in the similarity distributions of true positives and human-labelling errors (false positives), indicating a high similarity in their embeddings, which is notably larger than that observed with true negatives. In contrast, the overlap between true positives and synthetic label errors (false positives) is not obvious. This result demonstrates that human-labelling errors primarily arise due to high visual similarity between the assigned and latent class, unlike synthetic label errors. Additional results on other datasets and a quantitative evaluation are provided in the supplementary material.}

\zj{This empirical finding of a small overlap between true positives and synthetic label errors explains the effectiveness of synthetic noise-mitigating methods such as Sel-CL \cite{SSCL} and TCL \cite{huang2023twin}, which excel by allocating greater weight to confident pairs that are closely aligned. However, it also underscores the limitation of applying the same strategy to address the impact of human-labelling errors, which are close to true positives, serving as a primary motivation for this paper.}

\zj{Furthermore, the significant overlap between true positives and human-labelling errors reveals that human-labelling errors are `easy positives', indicating that the embeddings of positive samples are closely clustered in the representation space. This insight informs our strategy of reducing the weighting of easy positives as an effective means to mitigate the impact of false positives resulting from human-labelling errors.}

%

%

\subsection{\zj{Impacts of Human-Labelling Errors on SCL}}
\label{sec:analysis_on_truepair}
\zj{Based on the conclusions of Sec.~\ref{sec:similarity_of_mis}, we now know we need to focus on `easy positives'. It also raises the question: \textit{What is the impact of human-mislabelled samples when they appear as negatives?}} Therefore, in this section, we give a deeper analysis of the interaction of human-labelling errors and SCL by looking at the probability of false positives and false negatives during training. 


%

Let `A' represent an anchor image and `B' represent another paired image. If `A' and `B' are assigned the same label, they are a false positive if: `A' is correctly labeled while `B' is not (\zj{ \rom{1} in Fig.\ref{fig:CL_compare}(a)}); \textit{or} `A' is mislabelled while `B' is correctly labeled (\zj{ \rom{2} in Fig.\ref{fig:CL_compare}(a)}); \textit{or} `A' and `B' are mislabelled and do not 
belong to the same latent class (\zj{ \rom{3} in Fig.\ref{fig:CL_compare}(a)}). Conversely, if `A' and `B' are assigned different labels, they are a false negative if both `A' and 'B' are mislabelled but actually belong to the same latent class (\zj{ \rom{4} in Fig.\ref{fig:CL_compare}(a)}); \textit{or} `A' is correctly labeled and `B' is not, yet `B' belongs to the same latent class as `A' (\zj{ \rom{5} in Fig.\ref{fig:CL_compare}(a)}); \textit{or} `A' is mislabelled and `B' is not, with `A' being of the same latent class as `B' (\zj{ \rom{6} in Fig.\ref{fig:CL_compare}(a)}).

\zj{Assuming human-labelling errors rate is $\tau$, we can derive the probability of mislabelled data appearing as false negatives, $P_{FN} = \frac{\tau^2}{(C-2)^2} + \frac{2\tau - 2\tau^2}{C-1}$, and as false positives, $P_{FP} = 2\tau - \tau^2 - \frac{\tau^2}{(C-1)^2}$. $C$ is the number of classes and $\tau$ is the error rate. Since $\tau$ is small, terms with $\tau^2$ are negligible. Therefore, $P_{FN} \approx \frac{2\tau}{C-1}$ and $P_{FP} \approx 2\tau$. As $C$ increases, $P_{FN}$ tends to zero, while 
$P_{FP}$ remains constant. Therefore, with many classes and a small error rate, $P_{FP} \gg P_{FN}$. For example, if there are 200 classes in a dataset with a 5\% error rate, we would expect wrong learning signals from false positives and false negatives with probabilities of 9.75\% and 0.05\%, respectively.}

\zj{Additionally, we substantiate this finding with empirical evidence by quantifying the incorrect learning signals from human-labelling errors during training on the original CIFAR-100 and ImageNet-1K datasets separately, based on human-labelling errors provided by \cite{DBLP:conf/nips/NorthcuttAM21}.
When training on the CIFAR-100 dataset (comprising 100 classes), we found that 99.04\% of the incorrect learning signals were caused by human-labelling errors, from false positives, with only approximately 0.96\% stemming from negatives. Similarly, when training on the ImageNet-1K dataset (comprising 1000 classes), we discovered that 99.91\% of the incorrect learning signals were due to human-labelling errors, primarily from false positives, with only about 0.09\% arising from negatives.} We provide these rates for other datasets in the supplementary material.

\zj{Overall, both theoretical and empirical results show that when tackling labelling errors in contrastive learning, we can largely ignore false negatives due to their very low rate of occurrence, and focus on easy positives.
These observations motivate our proposed method, which incorporates less weighting on easy positives and reduces the wrong learning signal caused by human-labelling errors.}

%% file: sec/method.tex
\section{SCL with Robustness to Human-Labelling Errors}
\label{sec:method}
In this section, we describe our approach to \zj{mitigate} the impacts caused by human-labelling errors in positive pairs and how this fits into an overall contrastive learning objective. 
In Sec.~\ref{sec:label_analysis}, we noted that the most significant impact of the mislabeled samples arises when they are incorporated into the positive set and exhibit high similarity to the anchor (i.e.~easy positives). 
Our method therefore adheres to two key principles (Fig.~\ref{fig:CL_compare}b): (\textbf{P1}) it should ensure that the \textit{latent} class of positive samples matches that of the anchor~\cite{RN89,RN91,DBLP:journals/corr/abs-2007-00224,DBLP:journals/corr/abs-2010-01028}; and (\textbf{P2}) it should deprioritize easy positives, i.e.~those currently embedded near the anchor.
By reducing the weighting of easy positives, we minimize the effect of incorrect learning signals from false positive pairs. The model is also forced to recognize and encode deeper similarities that are not immediately apparent, improving its discriminative ability.

\subsection{Human-Labelling Errors in the SCL Objective}


Khosla~\textit{et al.}~\cite{RN81} argued that $\mathcal{L}_{\text{out}}^{\text{sup}}$ is superior to $\mathcal{L}_{\text{in}}^{\text{sup}}$, attributing this to the normalization factor $\frac{1}{|P(i)|}$ in $\mathcal{L}_{\text{out}}^{\text{sup}}$ that mitigates \zj{bias} within batches. Although $\mathcal{L}_{\text{in}}^{\text{sup}}$ incorporates the same factor, its placement inside the logarithm reduces its impact to a mere additive constant, not influencing the gradient and leaving the model more prone to \zj{this bias}.
We instead introduce a modified $\mathcal{L}_{\text{in}}^{\text{sup}}$ that directly reduces \zj{the impacts due to human-labelling errors} ,
and outperforms $\mathcal{L}_{\text{out}}^{\text{sup}}$.

We begin with a modified formulation of $\mathcal{L}_{\text{in}}^{\text{sup}}$ (Eq.~\ref{eq:supercon_in}), that is equivalent up to a constant scale and shift, but will prove easier to adapt:
%

\begin{equation}
\resizebox{0.68\textwidth}{!}{$
\mathbb{E}_{\substack{x\\ \left\{x_k^{+}\right\}_{k=1}^K \\\left\{x_i^{-}\right\}_{i=1}^N}}  \left[  \log \frac{-1}{\left |K  \right |}  \frac{\sum_{k=1}^{K}     e^{f(x)^T f(x_{k}^{+})}}{\sum_{k=1}^{K} e^{f(x)^T f( x_{k}^{+})}+\sum_{i=1}^N e^{f(x)^T f(x_i^{-})}}\right]
$}
\label{eq:our_supercon} 
\end{equation}

In \ref{eq:our_supercon}, all \( K \) samples from the same class within a mini-batch are treated as positive samples for the anchor $x$.

We now introduce our main technical contribution, which is devising an objective that mitigates \zj{human-labelling errors}, which consists of modifying Eq. (\ref{eq:our_supercon}).
As in Sec.~\ref{sec:label_analysis}, we assume there is set of latent classes \( \mathcal{C} \), that encapsulate the semantic content, and hopefully match the assigned labels.
%
Following \cite{DBLP:journals/corr/abs-1902-09229,DBLP:journals/corr/abs-2007-00224,DBLP:conf/iclr/RobinsonCSJ21}, pairs of images \( (x, x^{+}) \) are supposed to belong to the same latent class $c$, where $x \in \mathcal{X}$ is drawn from a data distribution $p(x)$.
Let $\tau$ denote the probability that any sample is mislabeled; we assume this is constant for all $x$. 
Since $\tau$ is unknown in practice, it must be treated as hyperparameter, or estimated based on previous studies.
We also introduce an (unknown) function \( z: \mathcal{X} \rightarrow \mathcal{C} \) that maps $x$ to its latent class label.
Then, \( p_{x}^{+} := p\big(x' \,|\, z(x') = z(x)\big) \) is the probability of observing \( x' \) as a positive example of \( x \), whereas \( p_{x}^{-} = p\big(x' \,|\, z(x') \neq  z(x)\big) \) is the probability of a negative example.
For each image \( x \), the objective (Eq.~\ref{eq:our_supercon}) aims to learn a representation \( f(x) \) by using positive examples \( \{x^{+}\}_{k=1}^{K} \)  with the same latent class label as \( x \) and negative examples \( \{x_{i}^{-}\}_{i=1}^{N} \) that belong to different latent classes.
Since $p$ is the true data distribution, the ideal loss function to be minimized if the latent labels $z(x)$ were known is:
%
%
\begin{equation}
\resizebox{0.68\textwidth}{!}{$
\mathcal{L}_{T}  = \mathbb{E}_{\substack{x\sim p \\ x_{k}^{+} \sim  p_{x}^{+} \\ x_{i}^{-} \sim  p_{x}^{-}}}   \left[   \frac{-1}{\left |K  \right |}       \log \frac{\frac{Q}{K}  \sum_{k=1}^{K}     e^{f(x)^T f(x_{k}^{+})}}{\frac{Q}{K} \sum_{k=1}^{K} e^{f(x)^T f( x_{k}^{+})}+ \frac{W}{N}\sum_{i=1}^N e^{f(x)^T f(x_i^{-})}}\right]
$}
\label{eq:supercon_weighted} 
\end{equation}
%
We term this loss function the \textit{true label loss}. Here, we have introduced weighting parameters \( Q \) and \( W \) to help with analysing the \zj{impacts of human-labelling errors}; when they equal the numbers of positive and negative examples respectively, $\mathcal{L}_T$ reduces to the conventional supervised contrastive loss (\ref{eq:our_supercon}). Note that supervised contrastive learning typically assumes \( p_{x}^{+} \) and \( p_{x}^{-} \) can be determined from human annotations (i.e.~$z(x)$ yields the \textit{assigned} label of $x$); however since we consider latent classes instead of assigned classes, we do \textit{not} have access to the true distribution.
However, we now show how to approximate this true distribution and improve the overall performance.

\subsection{\zj{Mitigating Human-Labelling Errors}}


For a given anchor $x$ and its embedding $f(x)$, we now aim to build a distribution $q$ on $\mathcal{X}$ that fulfils the principles \textbf{P1} and \textbf{P2}.
We draw a batch of positive samples \( \{ x_{k}^{+} \}_{k=1}^{K} \) from \( q \).
Ideally, we would draw samples from
%
\begin{equation}
    q^{+} (x^{+}) := \, q\big(x^{+}\,|\,z(x) = z(x^{+})\big) 
    \propto \, \frac{1}{e^{\beta f(x)^{T}f(x^{+})}} \cdot p_x^+(x^+)
    \label{eq:q}
\end{equation}
where $\beta \geq 0$. It is important to note that \( q^{+}(x^+) \) depends on \( x \), although this dependency is not explicitly shown in the notation.
The distribution is composed of two factors:
\begin{itemize}
    \item The event \( \{z(x) = z(x^{+}) \} \) indicates that pairs, \( (x, x^{+}) \), should originate from the same latent class (\textbf{P1}); recall $p^+_x(x^{+})$ is the true (unknown) positive distribution for anchor $x$.
    \item 
    The exponential term increases the probability of sampling hard positives, and decreases that of sampling easy positives (\textbf{P2}).
    This term is an unnormalized von Mises-Fisher density with mean direction \( f(x) \) and a concentration parameter \( \beta \). 
    The concentration parameter \( \beta \) modulates the weighting scheme of \( q^{+} \), specifically augmenting the weights of instances \( x^{+} \) that exhibit a lower inner product (i.e.~greater dissimilarity) to the anchor \( x \). 
\end{itemize}

The distribution \( q^{+} \) fulfils our desired principles of selecting true positives and deprioritizing easy positives.
However, we do not have the access to the latent classes, and so cannot directly sample from it.
We therefore rewrite it from the perspective of Positive-Unlabeled (PU) learning \cite{elkan2008learning,du2014analysis,DBLP:conf/iclr/RobinsonCSJ21,DBLP:journals/corr/abs-2007-00224}, which will allow us to implement an efficient sampling mechanism. 
We first define \( q^{-}(x^{+}) \propto \frac{1}{e^{\beta f(x)^{T}f(x^{+})}} \cdot p_x^-(x^+) \).
Then, by conditioning on the event \( \{ z(x) = z(x^{+}) \} \), we can write 
%
%
%
\begin{align}
    q (x^{+}) :=&\, \tau^{+}q^{+}(x^{+}) +\tau^{-}q^{-}(x^{+})
    \label{eq:pos_split}
\\ \Rightarrow \;
 q^{+}(x^{+})  =&\, \left( q (x^{+}) - \tau^{-}q^{-}(x^{+}) \right) / \tau^{+} 
 \label{eq:pos_rerrange}
\end{align}
where $\tau^{+}$ is the probability that a sample from the data distribution $p(x)$ will have the same latent class as $x$.  

We have now derived an alternative expression (\ref{eq:pos_rerrange}) for the positive sampling distribution \( q^{+} \) in terms of \( q \) and \( q^{-} \).   
Sampling directly from \( q \) and \( q^{-} \) is still not possible;
however, we can use importance sampling to approximate the necessary expectations. 
Specifically, we shall choose positives primarily by sampling an assigned-positive, while occasionally sampling an assigned-negative, with the probability of the latter set to counterbalance the mislabeling rate.

To achieve this we first consider a sufficiently large value for  \( K \) (i.e.~the number of positive samples for the anchor $x$) in the SCL objective (\ref{eq:supercon_weighted}), while holding the weighting parameter \( Q \) fixed. Then, (\ref{eq:supercon_weighted}) becomes:
%
%
\begin{equation}
\resizebox{0.68\textwidth}{!}{$
L_{T}  = \mathbb{E}_{\substack{x\sim p  \\ x^{-} \sim  p_{x}^{-}}}  \left[ \log \frac{-1}{\left |K  \right |}  \frac{Q  \mathbb{E}_{x^{+} \sim \zj{q^{+}}}\left[e^{f(x)^T f(x^{+})}\right] }{Q  \mathbb{E}_{x^{+} \sim \zj{q^{+}}}\left[e^{f(x)^T f\left(x^{+}\right)}\right] + \frac{W}{N}\sum_{i=1}^N e^{f(x)^T f(x_i^{-})}}\right]
    $}
\label{eq:positive_sampling} 
\end{equation}
%
%
By substituting Eq. \ref{eq:pos_rerrange} into Eq. \ref{eq:positive_sampling}, we obtain an objective that rectifies the impacts from human-labeling errors and also down-weights easy positives: 
%
%
%
%
%
\begin{equation}
\resizebox{0.68\textwidth}{!}{$
\mathbb{E}_{\substack{x\sim p \\ x^{-} \sim  q}}   \left[ \log \frac{-1}{\left |K  \right |}  \frac{\frac{Q}{\tau^{+}}\left(\mathbb{E}_{x^{+} \sim q}\left[e^{f(x)^T f(x^{+})}\right]-\tau^{-} \mathbb{E}_{v \sim q^{-}}\left[e^{f(x)^T f(v)}\right]\right)}    
{\frac{Q}{\tau^{+}}\left(\mathbb{E}_{x^{+} \sim q}\left[e^{f(x)^T f(x^{+})}\right]-\tau^{-} \mathbb{E}_{v \sim q^{-}}\left[e^{f(x)^T f(v)}\right]\right)   +  \frac{W}{N}\sum_{i=1}^N e^{f(x)^T f(x_i^{-})}}\right]
   $}
\label{eq:hard_pos}
\end{equation}
This suggests we only need to approximate the expectations $\mathbb{E}_{x^{+} \sim q}\left[e^{f(x)^T f\left(x^{+}\right)}\right]$ and $\mathbb{E}_{v \sim q^{-}}\left[e^{f(x)^T f(v)}\right]$ over \( q \) and \( q^{-} \), which can be achieved by classical Monte Carlo importance sampling, using samples from \( p \) and \( p^{-} \):
%
{
\fontsize{8}{8}\selectfont
\begin{align}
\mathbb{E}_{x^{+} \sim q}\left[e^{f(x)^T f\left(x^{+}\right)}\right] &= \mathbb{E}_{x^{+} \sim p}\left[e^{f(x)^T f(x^{+})} q / p\right]=\mathbb{E}_{x^{+} \sim p}\left[e^{(\beta+1) f(x)^T f(x^{+})} / Z(x)\right]
\label{eq:monte_pos}
%
\\
\mathbb{E}_{v \sim q^{-}}\left[e^{f(x)^T f(v)}\right] &= \mathbb{E}_{v \sim p^{-}}\left[e^{f(x)^T f(v)} q^{-} / p^{-}\right]=\mathbb{E}_{v \sim p^{-}}\left[e^{(\beta+1) f(x)^T f(v)} / Z^{-}(x) \right]
\label{eq:monte_pos_v}
\end{align}
}%
where \( Z(x) \) and \( Z^{-}(x) \) are the partition functions for \( q \) and \( q^{-} \) respectively. Hence, these expectations over \( p \) and \( p^{-} \) admit empirical estimates
%
\begin{equation}
    \widehat{Z}(x) =\frac{1}{M} \sum_{i=1}^M e^{\beta f(x)^{\top} f(x_i^{+})}
    \hspace{2em}\text{and}\hspace{2em}
    \widehat{Z}^{-}(x) =\frac{1}{N} \sum_{i=1}^N e^{\beta f(x)^{\top} f(x_i^{-})}.
\end{equation}


\paragraph{Mitigating label errors for negatives.}
%
%
Despite the minimal impact of mislabeled samples in negative sets (see Sec.~\ref{sec:label_analysis}), we extend our mitigation method to these samples to further reduce their adverse effects. Mirroring our strategy for positive samples, the mitigation process for negatives involves constructing a distribution that not only aligns with true negatives but also places greater emphasis on hard negatives.
We then use Monte Carlo importance sampling techniques to better estimate the true distribution of latent classes. Full details are given in the supplementary material.

\paragraph{\zj{Overall objective for SCL with robustness to human-labelling errors.}}
Using Eq.~\ref{eq:hard_pos} and incorporating mitigation for negatives, we get our final SCL-RHE loss
\begin{equation}\resizebox{1\textwidth}{!}{$
\mathbb{E}_{\substack{x\sim p\\ x^{+} \sim  q \\ x^{-} \sim  q}} \left[ \log \frac{-1}{\left |K  \right |}  \frac{\frac{Q}{\tau^{+}}\left(\mathbb{E}_{x^{+} \sim q}\left[e^{f(x)^T f(x^{+})}\right] - \tau^{-} \mathbb{E}_{v \sim q^{-}}\left[e^{f(x)^T f(v)}\right]\right)}{\frac{Q}{\tau^{+}}\left(\mathbb{E}_{x^{+} \sim q}\left[e^{f(x)^T f(x^{+})}\right] - \tau^{-} \mathbb{E}_{v \sim q^{-}}\left[e^{f(x)^T f(v)}\right]\right) + \frac{W}{\tau^{-}}\left(\mathbb{E}_{x^{-} \sim q}\left[e^{f(x)^T f(x^{+})}\right] - \tau^{+} \mathbb{E}_{b \sim q^{+} }\left[e^{f(x)^T f(b)}\right]\right)}\right]
$}
\label{eq:easy_contrastivelearning}
\end{equation}

 

%% file: sec/exp.tex
\section{Experiments}
\label{sec:exp}
We \zj{extensively} evaluate our proposed method, SCL-RHE, on image classification in three settings: training from scratch \zj{on datasets with human-labelling errors}, transfer learning using pre-trained weights \zj{(again with human-labelling errors)}, and pre-training on \zj{datasets with exceedingly high levels of synthetic label errors}. We also conduct several ablation experiments. For all experiments, we use the official train/test splits and report the mean Top-1 test accuracy across three distinct initializations.

We employ representative models from two categories of architectures -- BEiT-3/ViT base \cite{beit3,rn83}, and ResNet-50 \cite{RN36}.
While new state-of-the-art models are continuously emerging (e.g.~DINOv2 \cite{DBLP:journals/corr/abs-2304-07193}), our focus is not on the specific choice of architecture. Instead, we aim to show that SCL-RHE is model-agnostic and enhances performance using two very different architectures.
Further implementation details and the complete code for all experiments can be found in the supplementary material; the source code will be publicly available upon acceptance.

\subsection{Training from Scratch}
\label{subsec:results-pretrain}
\begin{table}[t]
\caption{Model accuracy measured using the acc@1 metric when trained with different loss functions on 3 popular image classification benchmarks. All models here are trained from scratch using only the indicated dataset, without pre-training.}
\label{table:scratch}
\centering
\resizebox{0.65\linewidth}{!}{
\scriptsize
\begin{tabular}{@{}llccc@{}}
\toprule
\textbf{Model} &\textbf{Loss} &\textbf{CIFAR-10}& \textbf{CIFAR-100}& \textbf{ImageNet-1K} \\
\midrule
\multirow{5}{*}{\textbf{BEiT-3}}&CE\cite{DBLP:journals/bstj/Shannon48}&71.70&59.67&77.91\\ 

                                &SupCon\cite{RN81}&88.96&60.77&82.57\\ 

                                &Sel-CL\cite{SSCL}&86.33&59.51&81.87\\ 

                                &TCL\cite{huang2023twin}&85.16&59.22&81.74\\ 

                                &Ours&\textbf{90.16}&\textbf{64.47}&\textbf{84.21}\\ 
\midrule
\multirow{5}{*}{\textbf{ResNet-50}}&CE\cite{DBLP:journals/bstj/Shannon48}&95.00&75.30&78.20\\ 

                                &SupCon\cite{RN81}&96.00&76.50&78.70\\ 

                                &Sel-CL\cite{SSCL}&93.10&74.29&77.85\\ 

                                &TCL\cite{huang2023twin}&92.80&74.14&77.17\\ 

                                &Ours&\textbf{96.39}&\textbf{77.82}&\textbf{79.15}\\ 
\bottomrule
\end{tabular}}

\end{table}


We first evaluate our proposed SCL-RHE objective in the pre-training setting, i.e.~training randomly initialized models from scratch without the use of additional data. For these experiments, we consider only human-labelling errors already present in the datasets without introducing synthetic errors. Following \cite{RN81}, to use the trained models for classification, we train a linear layer on top of the frozen trained models using a cross-entropy loss. 
We use three benchmarks: CIFAR-10, CIFAR-100 \cite{krizhevsky2009learning}, and ImageNet-1k \cite{DBLP:conf/cvpr/imagenet}.
Tab.~\ref{table:scratch} shows the performance of BEiT-3 and ResNet-50, with different loss functions on three popular image classification datasets. It is noteworthy that, due to the absence of pre-trained weights, BEiT-3 is identical to the ViT model \cite{beit3}. We compare against training with the standard cross-entropy loss and the state-of-the-art supervised contrastive learning loss (SupCon) \cite{RN81}. Additionally, we compare against two \zj{synthetic noise-mitigating} contrastive learning strategies (Sel-CL~\cite{SSCL} and TCL~\cite{huang2023twin}). We see that SCL-RHE consistently improves classification accuracy over other training objectives. On Imagenet-1k, SCL-RHE leads to a 6.3\% and 1.6\% improvement in accuracy for BEiT-3, relative to cross-entropy and SuperCon training, respectively. 

We find that SCL-RHE outperforms the existing contrastive methods Sel-CL and TCL (\zj{both designed to mitigate synthetic noise}), e.g.~on ImageNet-1K, SCL-RHE performs 2.4\% better than Sel-CL for BEiT-3 and 2.3\% better for ResNet50. 
We speculate that due to discarding many \zj{training} pairs, Sel-CL and TCL overfit a subset of training samples, limiting their performance in the realistic setting where the rate of label noise is relatively low (e.g. 5.85\% for CIFAR-100 \cite{DBLP:conf/nips/NorthcuttAM21}), and allowing them to be exceeded by SupCon. In contrast, SCL-RHE outperforms even SupCon, suggesting it is more applicable for real-world image training sets with low-to-moderate noise rates.

It is well known that transformer-based models underperform when training data is limited \cite{rn83,rn130,rn131}. This is highlighted by the low performance on CIFAR-10 and CIFAR-100 with cross-entropy training. We show that SCL-RHE, and to a lesser extent SuperCon, mitigate this---relative to cross-entropy, SCL-RHE gives a 19\% improvement on CIFAR-10. While BEiT-3 still fails to reach the performance of ResNet-50, the supervised contrastive approaches significantly close the gap, improving the applicability of transformer-based models in limited data scenarios. In the supplementary material, we also include an ablation study that measures the benefit of different aspects of our method. This shows that human-mislabelled samples impact the performance of supervised contrastive learning (SCL) due to their occurrence as soft positives, and that \zj{our proposed correction} on both positives and negatives helps to improve performance.

\begin{table}[t]
    \centering
    \begin{minipage}[t]{0.51\textwidth}
        \centering
        \caption{Accuracy of the BEiT-3 model using the metric acc@1 on different datasets and with various loss functions, when evaluating original and corrected test-set labels.}
        \label{table:model_accuracy}
        \resizebox{\textwidth}{!}{
        \begin{tabular}{@{}l|lcccccc@{}}
                \toprule
                \multicolumn{1}{@{}l}{\textbf{Loss}}& \multicolumn{1}{@{}l}{\textbf{Test set}}& \multicolumn{1}{l}{\textbf{CIFAR-10}} & \multicolumn{1}{l}{\textbf{CIFAR-100}} & \multicolumn{1}{l@{}}{\textbf{ImageNet-1K}} \\
                \midrule
                \multirow{2}{*}{\textbf{CE}\cite{DBLP:journals/bstj/Shannon48}}     & Original &  \multicolumn{1}{l}{71.70}& \multicolumn{1}{l}{59.67} & \multicolumn{1}{l}{77.91}\\
                       &                             Corrected  & 71.79 \texttt{\small\color{darkgreen}(+0.09)}& 59.82 \texttt{\small\color{darkgreen}(+0.15)} & 78.24 \texttt{\small\color{darkgreen}(+0.33)} \\ \midrule
                \multirow{2}{*}{\textbf{SupCon}\cite{RN81}}  & Original  &\multicolumn{1}{l}{ 88.96} &\multicolumn{1}{l}{60.77} &\multicolumn{1}{l}{82.57}   \\
                       &                              Corrected  & 89.11 \texttt{\small\color{darkgreen}(+0.15)} & 61.49 \texttt{\small\color{darkgreen}(+0.72)} & 83.74 \texttt{\small\color{darkgreen}(+1.17)} \\ \midrule
                \multirow{2}{*}{\textbf{Sel-CL}\cite{SSCL}}  & Original &\multicolumn{1}{l}{86.33} &\multicolumn{1}{l}{59.51} &\multicolumn{1}{l}{81.87}  \\
                       &                              Corrected  & 86.21 \texttt{\small\color{red}(-0.12)} & 58.62 \texttt{\small\color{red}(-0.89)} & 81.35 \texttt{\small\color{red}(-0.52)} \\ \midrule
                \multirow{2}{*}{\textbf{TCL}\cite{huang2023twin}}  & Original  &\multicolumn{1}{l}{85.16}  &\multicolumn{1}{l}{59.22}  &\multicolumn{1}{l}{81.74}  \\
                       &                             Corrected  & 84.97 \texttt{\small\color{red}(-0.19)} & 58.14 \texttt{\small\color{red}(-1.08)} & 81.28 \texttt{\small\color{red}(-0.46)} \\ \midrule
                \multirow{2}{*}{\textbf{Ours}}   & Original  &\multicolumn{1}{l}{90.16}  &\multicolumn{1}{l}{64.47} & \multicolumn{1}{l}{84.21} \\
                       &                              Corrected  &\textbf{90.41} \texttt{\small\color{darkgreen}(+0.25)} & \textbf{65.34} \texttt{\small\color{darkgreen}(+0.87)} & \textbf{86.02} \texttt{\small\color{darkgreen}(+1.81)}  \\ \bottomrule
        \end{tabular}}
    \end{minipage}%
    ~~
    \begin{minipage}[t]{0.48\textwidth}
        \centering
        \caption{Performance of ResNet-18 trained at different synthetic-noise levels. Time (Min/epoch) means the training time on a Nvidia A6000.}
        \label{table:noisy dataset}
        \resizebox{\textwidth}{!}{
        \begin{tabular}{@{}lcccccc@{}}
            \toprule
            \multicolumn{1}{c}{\textbf{Loss}}&\multicolumn{3}{@{}c}{\textbf{CIFAR-10}} &\multicolumn{2}{@{}c}{\textbf{CIFAR-100}}&\multirow{2}{*}{\textbf{Time}} \\
            \cmidrule(r){2-4} \cmidrule(l){5-6}
            \multicolumn{1}{@{}r}{\textit{Noise level}} & \textit{Original} & \textit{20\%}&\textit{40\%} & \textit{Original} & \textit{40\%}  \\
            \midrule
            \multicolumn{1}{@{}c}{\textbf{CE}\cite{DBLP:journals/bstj/Shannon48}} & 91.84 & 82.02 & 76.86 & 73.74 & 45.05 & \textbf{18.3} \\
            \multicolumn{1}{@{}c}{\textbf{SupCon}\cite{RN81}} & 94.08 & 89.13 & 79.57 & 74.58 & 51.33 & 20.2 \\
            \multicolumn{1}{@{}c}{\textbf{Sel-CL}\cite{SSCL}} & 91.42 & 94.45 & \textbf{93.22} & 72.10 & 74.24 & 29.1 \\
            \multicolumn{1}{@{}c}{\textbf{TCL}\cite{huang2023twin}} & 90.78 & 93.96 & 93.13 & 72.18 & \textbf{74.62} & 32.7 \\
            \multicolumn{1}{@{}c}{\textbf{Ours}} & \textbf{95.91} & \textbf{94.71} & 92.59 & \textbf{77.62} &74.08 & 20.4 \\  
            \bottomrule
        \end{tabular}%
        }
    \end{minipage}
\end{table}

\begin{table}[t]
\caption{Classification accuracy after fine-tuning a pre-trained BEiT-3 with different loss functions, on several benchmarks.}
\label{table:transfer}
\centering
\resizebox{\linewidth}{!}{
\begin{tabular}{@{}lccccccccc@{}}
\toprule
\textbf{Model} & \textbf{Loss} & \textbf{CIFAR-100} & \textbf{CUB-200} & \textbf{Caltech-256} & \textbf{Oxf-Flowers} & \textbf{Oxf-Pets} & \textbf{iNat2017} & \textbf{Places365} & \textbf{ImageNet-1K} \\ \midrule
ViT            & CE\cite{DBLP:journals/bstj/Shannon48}                 & 87.13              & 76.93            & 90.92               & 90.86                   & 93.81        & 65.26             & 54.06              & 77.91                \\
BEiT-3         & CE\cite{DBLP:journals/bstj/Shannon48}                 & 92.96              & 98.00               & 98.53               & 94.94                   & 94.49        & 72.31             & 59.81              & 85.40                 \\
BEiT-3         & SupCon\cite{RN81}           & 93.15              & 98.23            & 98.66               & 95.10                    & 94.52        & 72.85             & 60.31              & 85.47                \\
BEiT-3         & Sel-CL\cite{SSCL}           & 91.48              & 94.52            & 97.19               & 93.71                    & 94.51        & 72.43             & 58.36              & 85.21                \\
BEiT-3         & TCL\cite{huang2023twin}           & 90.92              & 93.89            & 97.26               & 93.89                    & 94.68        & 72.47             & 59.22              & 85.18                \\
BEiT-3         & Ours           & \textbf{93.81}              & \textbf{98.95}            & \textbf{99.41}               & \textbf{95.89}                   & \textbf{96.41}        & \textbf{76.25}             & \textbf{62.53}              & \textbf{86.51}                \\ \bottomrule
\end{tabular}}
\end{table}

\paragraph{Performance on corrected test sets.}
%
We next evaluate the same trained models, but using the corrected test-set labels from Northcutt \textit{et al.}~\cite{DBLP:conf/nips/NorthcuttAM21} (Tab.~\ref{table:model_accuracy}). \zj{With corrected test-sets, we reveal the true performance of trained models and the efficacy of various training methods.}
Importantly, we can observe a relatively larger increase in performance on the corrected test sets with SCL-RHE---e.g. an improvement of 1.81\% on ImageNet-1k.
This contrasts with SuperCon~\cite{RN81} and cross-entropy, which show a lesser improvement of 1.17\% and 0.33\%, respectively. \zj{This supports the claim that SCL-RHE is less prone to overfitting to human-labelling samples than SuperCon and cross-entropy, and the performance lead of SCL-RHE is larger than indicated by the noisy test sets.}
We find that Sel-CL\cite{SSCL} and TCL\cite{huang2023twin} generally lead to worse performance than SuperCon and do not show any performance gain when tested on the corrected labels. We speculate that, due to the relatively low mislabelling rates in these datasets (e.g.~5.85\% for ImageNet-1K), these approaches may be overly heavy-handed in combating labelling noise, diminishing the models' performance as a result. \zj{This confirms SCL-RHE achieves its design objective of addressing gaps in existing synthetic noise-mitigating methods by effectively rectifying human-labelling errors in widely used image training datasets without overfitting, thus yielding superior performance compared to these approaches.}  

\subsection{Transfer Learning}
\label{subsec:results-finetune}
We now assess performance when fine-tuning existing pre-trained models for specific downstream tasks.
Specifically, models are initialized using publicly available pretraining weights from ImageNet-21k \cite{DBLP:conf/cvpr/imagenet} and subsequently fine-tuned on smaller datasets, which retain their original human-labelling errors, using our proposed objective.
We use 8 datasets: CIFAR-100 \cite{krizhevsky2009learning}, CUB-200-2011 \cite{WahCUB_200_2011}, Caltech-256 \cite{caltech256}, Oxford 102 Flowers \cite{DBLP:conf/icvgip/flowers}, Oxford-IIIT Pets \cite{DBLP:conf/cvpr/pets}, iNaturalist 2017 \cite{DBLP:conf/cvpr/inat2017}, Places365 \cite{DBLP:conf/nips/places}, and ImageNet-1k \cite{DBLP:conf/cvpr/imagenet}. We select BEiT-3 base \cite{beit3} as the image encoder due to its excellent performance on ImageNet-1k.  Similar to \cite{RN81}, our approach for fine-tuning pre-trained models with contrastive learning involves initially training the models using a contrastive learning loss, followed by training a linear layer atop the frozen trained models using cross-entropy loss.


Tab.~\ref{table:transfer} shows classification accuracies after fine-tuning with different methods. 
We see that SCL-RHE gives the best classification accuracy across all datasets, with particularly large improvements on iNat2017 (+3.4\%) and state-of-the-art performance on Places365 (+2.2\%) when compared to fine-tuning with SupCon \cite{RN81}.
Similar to the pre-training setting, the synthetic noise-mitigating methods Sel-CL and TCL exhibit inferior performance compared to fine-tuning with cross-entropy and SCL-RHE across all 8 datasets. \zj{This provides further evidence that noise-mitigating methods designed for synthetic errors fail in addressing real human-labelling errors within the commonly adopted paradigm of pre-training followed by fine-tuning.
} 


\subsection{Efficiency Advantage and Robustness to Synthetic Noisy Labels}
\label{sec:robustness_noisy}

\zj{An important limitation of existing noise-mitigation methods is computational efficiency. They often add an extra module to measure the confidence of training samples \cite{huang2023twin} or calculate complicated graph relations to identify confident pairs \cite{SSCL}, which introduces extra computational costs. On the contrary, our proposed method has negligible computational overhead. We give the training time for our method and baselines in the rightmost column of Table~\ref{table:noisy dataset}. We observe that our SCL-RHE is much faster, reducing training time by 29.9\% compared to Sel-CL and 37.6\% compared to TCL.}

\zj{
Although our SCL-RHE aims at mitigating human-labelling errors in standard image datasets with typical error rates around 5\%, we also tested it for resilience against higher synthetic noise rates ($\geq$20\%), far exceeding normal benchmark conditions. We trained models from scratch on CIFAR datasets, randomly noised by altering a portion of labels to different classes, thereby challenging our initial focus on natural human errors and the findings that human-mislabeled samples are visually similar to their assigned class. By employing ResNet-18  to match the setting of \cite{huang2023twin,SSCL} and comparing SCL-RHE against techniques like CE, SupCon (a state-of-the-art general SCL method), and noise-mitigating SCL methods Sel-CL and TCL, we evaluate our method's performance across varying noise levels, including the original dataset, 20\%, and 40\% synthetic noise additions. Results in Table~\ref{table:noisy dataset} reveal that SCL-RHE outperforms these methods in handling both the original dataset and its 20\% noised variant. Even under an extreme 40\% error rate, SCL-RHE remains competitive, slightly trailing behind Sel-CL and TCL on CIFAR-100 by 0.16\% and 0.54\% in accuracy respectively, yet significantly outperforming CE and SupCon. This underscores SCL-RHE's capability to effectively mitigate not only realistic but also to withstand a degree of synthetic label errors.
}

%% file: sec/conclusion.tex
\section{Discussion and Conclusion}
\paragraph{Limitations.}
%
Although SCL-RHE exceeds the state-of-the-art, it still has certain limitations.
First, while existing research has estimated mislabelling rates for various datasets \cite{DBLP:conf/nips/NorthcuttAM21}, determining this for new datasets remains a challenge.
This issue could be effectively managed by adopting the typical average error rate of $3.3\%$, as reported in \cite{DBLP:conf/nips/NorthcuttAM21}, as a baseline for hyperparameter tuning to identify an optimal value. Moreover, in our experiments, we observed that SCL-RHE's performance exhibits low sensitivity to the estimated mislabelling rates, with further details of this study provided in the supplementary materials. 
\zj{Second, although our SCL-RHE outperforms existing methods even for mitigating synthetic errors up to 20\% noise rates, it does not surpass Sel-CL \cite{SSCL} and TCL \cite{huang2023twin} in scenarios involving extremely high synthetic error rates of 40\%. This is because our model is tailored to mitigate human-labelling errors in datasets with error characteristics typical of real-world scenarios—i.e.~where mislabellings occur naturally due to human error when classes are genuinely similar or ambiguous.
}


\paragraph{Conclusion.}
 \zj{In this work, we investigated the extent and manner in which human-labelling errors impact supervised contrastive learning (SCL), and demonstrated these impacts diverge from those on regular supervised learning. Based on this, we introduced a novel SCL objective that is robust to human errors, SCL-RHE, specifically designed to mitigate the influence of real human-labelling errors (instead of synthetic noise addressed in previous works). Our empirical results reveal that SCL-RHE consistently outperforms traditional cross-entropy methods, the previous state-of-the-art SCL objectives, and noise-mitigating approaches designed for synthetic noise, both when training from scratch and in transfer learning. In addition to its superior performance, a key advantage of SCL-RHE is its efficiency---unlike previous methods that mitigate synthetic label noise, it incurs no extra overhead during training.}
